# How Generative-AI can be Effectively used in Government Chatbots ——A Joint Experimental Research based on Large Language Model

Zeteng Lin
Data Science and Analytics Thrust
zlin810@connect.hkust-gz.edu.cn

**Abstract**

*With the rapid development of artificial intelligence and breakthroughs in machine learning and natural language processing, intelligent question-answering robots have become widely used in government affairs. This paper conducts a horizontal comparison between Guangdong Province's government chatbots, ChatGPT, and Wenxin Ernie, two large language models, to analyze the strengths and weaknesses of existing government chatbots and AIGC technology. The study finds significant differences between government chatbots and large language models. China's government chatbots are still in an exploratory stage and have a gap to close to achieve "intelligence." To explore the future direction of government chatbots more deeply, this research proposes targeted optimization paths to help generative AI be effectively applied in government chatbot conversations.*

## 1. Introduction

Generative Artificial Intelligence (abbreviated as "AIGC" for AI Generative Content) refers to a production method that uses artificial intelligence to automatically generate content. AIGC tools utilize deep learning methods to learn existing knowledge, and when prompted, generative artificial intelligence can create new content based on statistical models.[1] In constructing artificial neural networks, AIGC employs supervised, semi-supervised, and unsupervised learning methods to process both labeled and unlabeled data, enabling the generation of various types of content such as text, code, images, audio, and video. The emergence of AIGC tools, exemplified by ChatGPT, has sparked profound discussions worldwide, signifying a global shift in intelligent technology towards a more creative direction.[2]

At the end of November 2022, the American company OpenAI launched an artificial intelligence dialogue chatbot, ChatGPT.[3] The launch of ChatGPT sparked a wave of interest in Large Language Models (LLMs). AIGC has been rapidly adopted in various fields. In the realm of public administration, governments worldwide have integrated AIGC into their service response platforms on government websites. This integration aims to facilitate rapid and personalized responses, significantly reducing the workload of public service personnel. In the United States, Arizona's Department of Economic Security has deployed a chatbot named "Dave" to assist citizens in understanding public finance information.[4] Similarly, the Maryland Department of Labor utilizes "Dayne", a chatbot providing information related to unemployment benefits.[5] Finland has introduced "Kamu", a chatbot aimed at offering citizen services, demonstrating the broad applicability of these technologies in various administrative functions.[6] In the United Kingdom, the National Health Service (NHS) has integrated AIGC into its chatbot, "Ada Health", to offer health-related advice, including vaccine information and diagnosis, thus facilitating personalized health assessments.[7] Moreover, Estonia has launched a virtual assistant called "Suve", designed to provide accurate and reliable information in response to public inquiries.[8] These instances highlight the global trend of governments seeking more efficient response models through the adoption of advanced AI chatbot technologies.

This study investigates Guangdong Province's government chatbots, ChatGPT, and Wenxin Yiyi, two LLMs, by posing various administrative questions of different aspects and complexities. Utilizing text analysis, index evaluation, joint experimentation, and Natural Language Processing (NLP) technologies, the study delves deeply into exploring these two LLMs and gathering data. This paper inputs procedural and complex questions into both ChatGPT and Wenxin Yiyi, conducting preliminary analyses of the responses, including similarity, sentiment, word frequency, and word vector analyses.

This experiment seeks to understand public preferences for the development of chatbots, further aiding the effective application of generative AI in government chatbots.

## 2. Literature Review

The integration of AI in government affairs is a rapidly evolving area, with many governments exploring AIGC, tools like ChatGPT, and chatbots to enhance public services. We synthesize current knowledge and applications in this field, focusing on the implications, challenges, and real-world examples.

Governments are increasingly turning to AI for smarter policymaking, reimagined service delivery, and more efficient operations. AI and data analytics can enhance the effectiveness of policymaking, providing decision-makers with tools to deliver more value. For example, AI can analyze demographic trends to help policymakers identify issues and intervene with informed policies. Additionally, NLP can eliminate policy conflicts prior to implementation, and social media sensing can gauge constituent sentiment.[9] In terms of service delivery, AI can revolutionize how citizens interact with government services, such as using job seeker data to optimize employment agency support or managing health system resources for efficiency and cost reduction. An AI-powered future could involve integrated service delivery, where a single point of entry connects citizens to all relevant government services, tailored to individual needs.

ChatGPT, an AI model developed by OpenAI, holds promise for automating tasks and increasing productivity in government. In practical applications, ChatGPT can serve as an interactive tool to answer resident queries, assist in documentation, engage citizens, explain policies, aid in training and onboarding, handle emergency communications, improve accessibility for people with disabilities, and offer multilingual support. However, it's vital to ensure that ChatGPT is trained with accurate, reliable sources, particularly for sensitive government information.[10] There is also a work from Carvalho[11] that proposes the potential of ChatBots in public management affairs from the perspective of the relationship between Portuguese local government decision-makers and citizens, thereby optimizing and improving solutions for the cooperative relationship between citizens and municipalities.

However, there are still difficulties in the application and management of AIGC. Algorithm bias, data abuse, privacy threats, technological monopoly caused by differences in the ability to hold the technology, and obscured intellectual property rights in the generated content are all possible risks in the application process of AIGC.[12] Another work proposed that it is necessary to use public power to restrict GPT applications at the government level, especially at the regulatory level. It is difficult to avoid the harmful content that AIGC may generate, including false information and illegal information, by relying solely on R&D companies to regulate operating procedures. On this basis, the government should build and operate an ecosystem around the "Gov+GPT" model, exert its regulatory functions, support companies from the data level in screening the source of AI, and exert its backbone.[13]

People's trust in chatbots needs deeper exploration. A work from Ma[14] proposed the AIDUA model, combined with cognitive evaluation theory, to evaluate the acceptance of ChatGPT from multiple dimensions such as society, human nature, value, and novelty. evaluation, proposed using perceived usability and perceived ease of use as mediators to enhance public trust. Another social survey of Chinese citizens showed that their preference for chatbots is more reflected in intelligence, initiative, responsibility, etc. The relative importance of the social characteristics reflected by them provides powerful information communication services for chatbots.[15]

In summary, scholars have conducted extensive research on AIGC and government chatbots. Regarding AIGC, some studies have provided definitions and interpretations of its development. Additionally, some scholars have explored governance strategies for AIGC from various perspectives, including legal regulations, technical adjustments, multi-party collaboration, and ethical norms. They have also proposed technical governance solutions. Regarding government chatbots, research primarily focuses on two aspects. One aspect examines the factors influencing government adoption of chatbots, while the other investigates the factors affecting users' willingness to use and satisfaction with government chatbots. However, there are some shortcomings in the existing research. Firstly, most domestic studies on government chatbots are limited to empirical analysis and policy analysis, lacking diversity in research perspectives. Secondly, the research often focuses on domestic government chatbots and does not compare them with international AIGC models. Furthermore, there is a lack of research on how AIGC can be integrated with government chatbots in the domestic context.

This research builds upon the work of leading scholars in the field and introduces innovations. Firstly, there is innovation in the choice of research subjects. This study tests two typical LLMs, namely, the domestic *Ernie* and the foreign ChatGPT. Through multiple tests on the same set of questions, a comparative analysis is conducted on the strengths and weaknesses of these two LLMs based on five key characteristics defined in the research design. Secondly, there is innovation in the research methods. This study involves a preliminary textual analysis of the responses generated by the two models, including similarity analysis, word frequency analysis, responsibility analysis, communicative analysis, and user-friendliness analysis. Building upon this textual analysis, a joint experiment approach is employed, primarily relying on public ratings. The objective of this research is to explore the possibility of incorporating the advantages of AIGC into government chatbots through comparative experiments. It aims to propose optimization strategies so that the integration of AIGC can help government question-answering chatbots better handle issues such as unclear user expressions and non-standard questions.

## 3. Methodology

The main purpose of this article is to explore whether the existing AIGC technology can be effectively applied to robot government affairs chat and propose feasible optimization directions. This article references the "Personal Affairs" section of the National Government Service Website, selects 5 consultation areas based on the span of life, and divides them into two levels: procedural problems and complex problems. Test questions are designed based on the keywords of each service section and continuous follow-up questions are conducted (shown as in Table 1, Table 2).

Table 1 Summary of procedural issues

| Consulting field | Specific problem | Ask |
|---|---|---|
| Education | How to report illegal compensation by off-campus education institutions class? | 1. What materials are needed to report illegal make-up classes at off-campus institutions?<br>2. How long does it take for the result to be announced after a report is made?<br>3. Will the whistleblower's information be disclosed when filing a report? |
| Employment and entrepreneurship | How to register a personal online store in China | 1. Do I need a business license to register a personal online store?<br>2. Are there any entrepreneurial policy subsidies after applying for a business license?<br>3. Is there any fee required to register a personal online store? |
| Social security and medical care | How to use medical insurance for reimbursement? | 1. How to determine if my situation falls within the medical insurance reimbursement conditions of this city?<br>2. After using medical insurance for reimbursement, how long does it take for the money to arrive?<br>3. Can I use medical insurance when I am in a different place? |
| Public security | How to proceed with the household registration transfer procedure? | 1. What documents are required?<br>2. Do I need to make an appointment?<br>3. Which institution or platform should I go to make an appointment? |
| Retirement care | Can I apply for a pension for the elderly at home? | 1. What is the pension application process?<br>2. How do I know the amount of pension I can apply for?<br>3. If the elderly at home have difficulty moving, can I apply on behalf of the elderly at home? |

Table 2 Summary of complexity issues

| Specific problem | Ask |
|---|---|
| I received a notice that my house is about to be demolished by the government. What should I do? | 1. How to confirm whether the demolition of a house is legal?<br>2. Can I appeal?<br>3. Forced demolition involves subsequent accommodation problems. How to solve the problem? |
| There is a flood near my home and my home is flooded. What should I do? | 1. Can I get subsidies from relevant agencies?<br>2. Does the city have emergency relief and resettlement measures?<br>3. I felt so helpless, what was I going to do with my flooded home? |
| The newly built building near my home is still being renovated after ten o'clock in the evening. I have contacted the property management company there several times to no avail. What should I do? | 1. Is there any law that can define whether their actions violate the rights of residents?<br>2. How else can I reflect the channel?<br>3 I'm very angry and plan to teach the property management there a lesson. Do you have any good suggestions? |
| There is a garbage dump near my school. When the garbage is burned, there is a foul smell, which is very annoying. What should I do? | 1. The health of many faculty and staff has been affected. How should I report this issue to the relevant departments?<br>2. What protective measures should I take?<br>3. A garbage incineration plant is built near a school without permission. Can I complain? |
| I am a property owner. After purchasing a new house, what should I do if I encounter an unfinished building and feel very stressed? | 1. I feel anxious and panicked every day because of this problem. Should I contact a psychiatrist?<br>2. Although I have encountered an unfinished building, I still need to repay the mortgage. Can I appeal not to repay it?<br>3. Disputes over unfinished buildings may involve legal proceedings. How should I seek legal assistance? |

This article initially inputs procedural problems and complex problems into the Guangdong Government Affairs Chatbot, as well as ChatGPT and Wenyi One-sentence, two LLM models, to obtain corresponding results and conduct text analysis. However, during the preliminary testing, the response of the Guangdong Government Affairs Chatbot often provides links directly, resulting in vague and general answers, making it difficult to provide satisfactory responses to public government affairs questions. Therefore, this article focuses on studying how to apply AIGC to government affairs chatbots, fill in the deficiencies of government affairs chatbots, and make responses more in line with public demands. Preliminary text analysis is conducted based on the text data generated from AIGC answers, including topic analysis, similarity analysis, responsibility analysis, and sentiment analysis.

In order to better analyze the application of artificial intelligence in the field of governance, we extract the text results, text representation, and their features for text mining. This article applies the following text analysis techniques to analyze artificial intelligence texts:

Topic Analysis: Through topic analysis, we discover hidden and meaningful information in the text collection. Algorithms such as Latent Dirichlet Allocation (LDA) are used to identify the topics in the text and classify the text accordingly. This helps us discuss the main issues or trends in our text analysis.

Similarity Analysis: Similarity analysis is used to compare the similarity between texts. We use it to calculate the cosine similarity or Jaccard similarity coefficient between texts to discover the similarity and differences between policy documents or legislative texts. Word vector analysis is one type of similarity analysis. It is a technique derived from Natural Language Processing (NLP) that converts vocabulary into numerical vectors to capture the semantic information of the vocabulary. Models such as Word2Vec or GloVe are used to discover semantically similar vocabulary, perform word disambiguation, and explore the relationships between vocabulary. This helps us explore the relationships between concepts and topics in policy texts, especially between different responses.

Sentence Quantification: Sentence quantification evaluates the information density and complexity of a text by analyzing the number and structure of sentences. We use it to assess the readability and information load of policy documents.

Average Sentence Depth: Average sentence depth is usually determined through syntactic analysis and measures the complexity of sentence structure. In a syntactic tree, depth refers to the longest path from the root to a leaf (lexical item). Average depth can reflect the language complexity of the text and the complexity of sentence construction. We use it to evaluate the comprehensibility and formal rigor of the text, which is crucial for ensuring clear communication of information.

Sentiment Analysis: Sentiment analysis is used to understand the emotional state of artificial intelligence by identifying the sentiment tendencies in the text. In the field of governance, it can be used to measure public sentiment towards policies or political events, thereby assessing the sentiment orientation of artificial intelligence. In this article, sentiment analysis is mainly used to determine the user-friendliness and communicability of chatbots.

Through the aforementioned text analysis techniques, we are able to delve into text data, uncover the structure and content of artificial intelligence documents, and identify the core issues and specific concerns in policy texts. This allows for a better understanding of how artificial intelligence is discussed and applied in the field of governance. These analyses not only reveal high-frequency keywords and topics, but also uncover the deeper meanings and public sentiments in the text, providing data support for policy-making and public services.

## 4. Research Results

Recently, the intelligence and informatization of government services have garnered increasing attention from the public. How to use advanced technology to enhance the quality and efficiency of government services has become a crucial topic in government innovation. ChatGPT and Ernie, as prominent examples of AIGC, are attracting more attention for their potential applications in government Q&A.

This section aims to verify the accuracy and usefulness of ChatGPT and Ernie in handling government-related questions, providing decision support and references for government departments to integrate AI into digital services. Detailed textual data can be found in Appendices 2 and 3.

### 4.1. Topic Analysis

Topic analysis, commonly used in the fields of text mining, information retrieval, information organization, content recommendation, and user intent recognition, is also often referred to as topic modeling. This technique is utilized for automatically identifying potential topic content from a large volume of documents. The key idea of topic analysis is that each document in a document collection is composed of hidden topics that determine the distribution of words. Through statistical learning methods, it is possible to infer the topic distribution of documents and the relationships between topics and words.

Major topic analysis techniques include Latent Semantic Analysis (LSA), Latent Dirichlet Allocation (LDA), Non-negative Matrix Factorization (NMF), etc. In this paper, considering the challenges associated with unclear distinctions resulting from too many topics and insufficient textual data with too few topics, we use perplexity as a measure to determine the number of topics.

Figure 1 illustrates the results of topic analysis using the Latent Dirichlet Allocation (LDA) method, including topics such as environment, community reporting, government support, health insurance policies, document applications, and emotional anxiety. Furthermore, chatbot text is clearly informative and instructive, with both Ernie and ChatGPT having a structured response style.

### 4.2. Similarity Analysis

Similarity analysis is a crucial method for in-depth study of different textual content features and distinctions,

particularly in AI fields where semantic text similarity algorithms support tasks such as question answer systems and intelligent retrieval.

This paper conducts a logic and text of Ernie and ChatGPT's QA and analyses their similarity, and investigates the responsiveness and matching of different models, thereby providing a direction for subsequent research.

Cosine similarity is a commonly used text similarity method, calculating similarity by measuring the cosine angle between two vectors. The specific process involves: 1. Converting text into vectors through methods such as TF-IDF or Word Embeddings. 2. Comparing the similarity of two sets by Jaccard similarity metric. 3. Calculating text similarity by the Longest Common Subsequence (LCS).

However, similarity analysis requires pre-training data, and to scientifically compare the similarity of two models, we choose the pre-trained BERT (Bidirectional Encoder Representations from Transformers) model to ensure experimental reproducibility. The BERT model, combined with a mask mechanism and an added the prediction "next sentence prediction" task, better understands the context of the text and generates high-dimensional vectors that represent the semantics of the text.

Table 3 presents a similarity analysis using three different methods: BERT similarity, TF-IDF similarity, and Word Embeddings similarity. The analysis includes the mean, standard deviation (Std), minimum (Min), and maximum (Max) values for each method. BERT similarity shows a high average score of 0.816, indicating a strong level of similarity between the compared elements, with a relatively low standard deviation of 0.055, suggesting consistent results. The range for BERT similarity is between 0.698 and 0.936.

In contrast, TF-IDF similarity has a much lower mean (0.011), indicating a generally low level of similarity, and this is further emphasized by its wide range from 0.000 to 0.368. The standard deviation for TF-IDF is also 0.055, showing a moderate level of variability in the results. Word Embeddings similarity has a mean of 0.016, similar to TF-IDF, but with a significantly higher standard deviation of 0.132, indicating a wider variation in similarity scores. Its range is the broadest, from -0.226 to 0.675, reflecting the potential for both high dissimilarity and similarity. This analysis suggests that while BERT offers more consistently high similarity scores, TF-IDF and Word Embeddings methods exhibit greater variability and generally lower levels of similarity.

Table 3 Similarity Analysis

|      | BERT similarity | TF-IDF similarity | Word Embeddings similarity |
| --- | --- | --- | --- |
| Mean | 0.816 | 0.011 | 0.016 |
| Std  | 0.055 | 0.055 | 0.132 |
| Min  | 0.698 | 0.000 | -0.226 |
| Max  | 0.936 | 0.368 | 0.675 |

### 4.3. Conscientiousness Analysis

Conscientiousness refers to the sense of responsibility and meticulousness exhibited by a chatbot when responding to user requests. In this paper, we hypothesize that if a chatbot can respond more fully and accurately to text when responding to a question, the higher conscientiousness is able to express more complex relationships and concepts in its responses, indicating that responses with conscientious bot encompass more subordinate clauses or other nested structures.

In this paper, spaCy, a dependency parsing, and syntactic analysis tool derived from natural language processing, is used to understand the structure of the sentence, which helps this paper to understand the relationship between the words in the sentence.

First, the number of sentences is counted, reflecting the information density and structural complexity of the text.

Firstly, the study calculates the number of sentences, reflecting the information density and structural complexity of the text.

Secondly, it analyzes the average depth of dependency relationships between words in sentences. Responses with more subordinate clauses and nested structures, contributing to greater grammatical complexity, are considered more proactive and conscientious.

Thirdly, the study measures the complexity of sentence grammar, defining words with a length greater than 6 as complex words, and defining text complexity as the ratio of complex words to the total number of words.

Table 4's conscientiousness analysis comparing Ernie and ChatGPT shows distinct patterns in their communication styles. Ernie tends to provide responses with greater depth, as indicated by a higher average depth and a wider range, yet with fewer sentences, suggesting a concise but layered approach. ChatGPT, in contrast, generally uses more sentences per response and exhibits a slightly higher complexity in word choice, pointing to a tendency for longer and more varied responses. The standard deviation values highlight this contrast further, with Ernie showing more consistency in depth and sentence number, while ChatGPT demonstrates greater variability across these metrics, reflecting a more diverse

range of responses in terms of length and complexity.

Table 4 Conscientiousness Analysis

|  | Ernie | | | ChatGPT | | |
|---|---|---|---|---|---|---|
|  | Depth | Sentences Number | Complex Word | Depth | Sentences Number | Complex Word |
| Mean | 2.26 | 1.14 | 0.4374 | 1.12 | 1.4 | 0.4720 |
| Std. | 1.860 | 0.639 | 0.0807 | 0.3854 | 1.38505 | 0.0850 |
| Min | 1 | 1 | 0.25 | 1 | 1 | 0.1875 |
| Max | 7 | 5 | 0.5 | 3 | 7 | 0.5 |
| N | 50 | 50 | 50 | 50 | 50 | 50 |

## 4.4. Sentiment Analysis

Chatbots conveying emotions to the public may have an impact on the communication and subsequently influence the public's affinity, trust, and satisfaction levels towards the machine. Therefore, this paper develops a sentiment analysis aiming to discuss the role of chatbot's emotional expression in governmental communication. The specific steps are as follows:
1. Preprocess the text data, including Chinese word splitting and stop-word removal, to ensure that the text input to the model is clean and formatted.
2. Utilize SnowNLP in Python for sentiment analysis, quantifying the emotional tendency with probability values as sentiment scores. A score closer to 0 indicates a more negative sentiment conveyed by the text, while a score closer to 1 signifies a more positive emotional expression. Observing the sentiment distribution charts, it can be noted that Yiyian exhibits a more balanced emotional expression, whereas ChatGPT's sentiment expression demonstrates a more pronounced trend of polarization. This suggests that compared to Ernie, ChatGPT's emotional tone is more distinct, and the peak of positive emotions implies that ChatGPT may tend to use encouraging or positive language in generating responses. Both chatbots exhibit the generation of encouraging language or expression of positive emotions in their responses. This phenomenon aids chatbots in leveraging their mediation and communication capabilities in human-machine interactions, enhancing public affinity, trust, and satisfaction when providing governmental services.

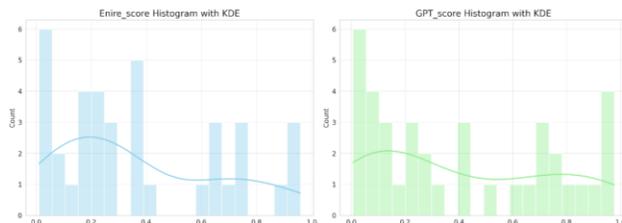

Figure 1 Sentiment Analysis between Ernie and GPT

## 5. Conclusion and Recommendations

With the continuous advancement of artificial intelligence, various local governments in China have experimented with the application of government service robots to assist in administrative consultations. However, the Q&A robots in government websites has remained stagnant in the image of "search engine" with dialogue boxes, resulting in limited public interest and user experience. The recent popularity of interactive models such as ChatGPT and Ernie has broken people's stereotypical impressions of machine conversations, with AIGC becoming a crucial tool for white-collar professionals. Therefore, how to apply AIGC to enhance the quality and efficiency of government services has become a pivotal topic in governmental innovation.

This research, based on text analysis, conducted a comprehensive analysis of the response texts from ChatGPT and Ernie in four dimensions: topic analysis, similarity analysis, conscientiousness analysis, and sentiment analysis. In addressing responses to complex questions, both models exhibited specific response styles, including the use of encouraging language and expressions of positive emotions. Using ChatGPT as a reference, Ernie's responses demonstrated higher conscientiousness and a more balanced expression of rational emotions.

This study holds significant implications for promoting the application and dissemination of intelligent Q&A robots in governmental services. By introducing three major scoring dimensions—conscientiousness, communicativeness, and populism—and comparing the overall performance of the two AIGC models, the study elucidates public preferences in governmental conversations, indicating optimization directions for government robots. For example, providing detailed, precise, and effective solutions to public inquiries, genuinely addressing the questions posed, and alleviating the burden of manual Q&A. Additionally, it should focus on responding to the emotional needs of inquirers, and use encouraging language to mitigate negative emotions.

These research findings provide references for the future development of intelligent Q&A robots.